\documentclass[10pt,twocolumn,letterpaper]{article}

\usepackage{iccv}

\usepackage{epsfig}
\usepackage{graphicx}
\usepackage[dvipsnames]{xcolor}

\usepackage{times}
\usepackage{amsmath}
\usepackage{amssymb}

\usepackage{booktabs}
\usepackage{multirow}

\usepackage{authblk}
\renewcommand*{\Affilfont}{\normalsize\normalfont}
\makeatletter
\renewcommand\AB@affilsepx{, \protect\Affilfont}
\makeatother

\newcommand{\one}{({\em i}\/)\xspace}
\newcommand{\two}{({\em ii}\/)\xspace}

\definecolor{TableGray}{gray}{0.35}
\definecolor{RoyalBlue}{RGB}{0,113,188}
\definecolor{PipelineGreen}{RGB}{59,181,75} 
\definecolor{PipelineRed}{RGB}{237,94,95} 
\definecolor{PipelineBlue}{RGB}{32,100,155} 

\usepackage[pagebackref=true,breaklinks=true,letterpaper=true,colorlinks,bookmarks=false]{hyperref}

\iccvfinalcopy 

\def\iccvPaperID{****} 

\ificcvfinal\fi

\begin{document}

\title{Seeing and Hearing Egocentric Actions: How Much Can We Learn?}

\author[1]{Alejandro Cartas}
\author[2]{Jordi Luque}
\author[1]{Petia Radeva}
\author[2]{Carlos Segura}
\author[3]{Mariella Dimiccoli}
\affil[1]{University of Barcelona}
\affil[2]{Telefonica I+D, Research, Spain}
\affil[3]{Institut de Rob\`otica i Inform\`atica Industrial (CSIC-UPC)}

\renewcommand\Authands{ and }

\maketitle

\newcommand{\pt}[1]{\textcolor{red}{PT: #1}}

\begin{abstract}
Our interaction with the world is an inherently multimodal experience. However, the understanding of human-to-object interactions has historically been addressed focusing on a single modality. In particular, a limited number of works have considered to integrate the visual and audio modalities for this purpose. In this work, we propose a multimodal approach for egocentric action recognition in a kitchen environment that relies on audio and visual information. Our model combines a sparse temporal sampling strategy with a late fusion of audio, spatial, and temporal streams. Experimental results on the EPIC-Kitchens dataset show that multimodal integration leads to better performance than unimodal approaches. In particular, we achieved a $5.18\%$ improvement over the state of the art on verb classification.

\end{abstract}

\section{Introduction}
The ability to integrate multisensory information is a fundamental feature of the human brain that allows efficient interaction with the environment \cite{ernst2004merging}. To mimic this human characteristic is crucial for autonomous robotics to reduce ambiguity about sensory environment and to form robust and meaningful representations. Beside artificial agents, multimodal integration has been used in traditionally vision-based tasks such as scene classification \cite{aytar2016soundnet,zhang2001audio}, social behaviour analysis \cite{alghowinem2016multimodal,guccluturk2016deep,pantic2007human} and activity recognition \cite{Arandjelovic17,Owens2016VisuallyIS}. While scene classification \cite{aytar2016soundnet,zhang2001audio} and social behaviour analysis \cite{alghowinem2016multimodal,guccluturk2016deep,pantic2007human} have been approached by integrating mostly audio and visual features, comparative little attention has been payed to audiovisual integration for activity recognition \cite{Arandjelovic17,Owens2016VisuallyIS,subedar2019uncertainty}. Indeed, most existing approaches for the latter aimed at combining different inertial sensors such as accelerometers, gyroscopes and magnetic field sensors \cite{ordonez2016deep} or inertial cues with audio or depth cues \cite{bulling2014tutorial,chen2017survey}. 

\begin{figure}[!t]
\begin{center}
\includegraphics[width=0.85\columnwidth]{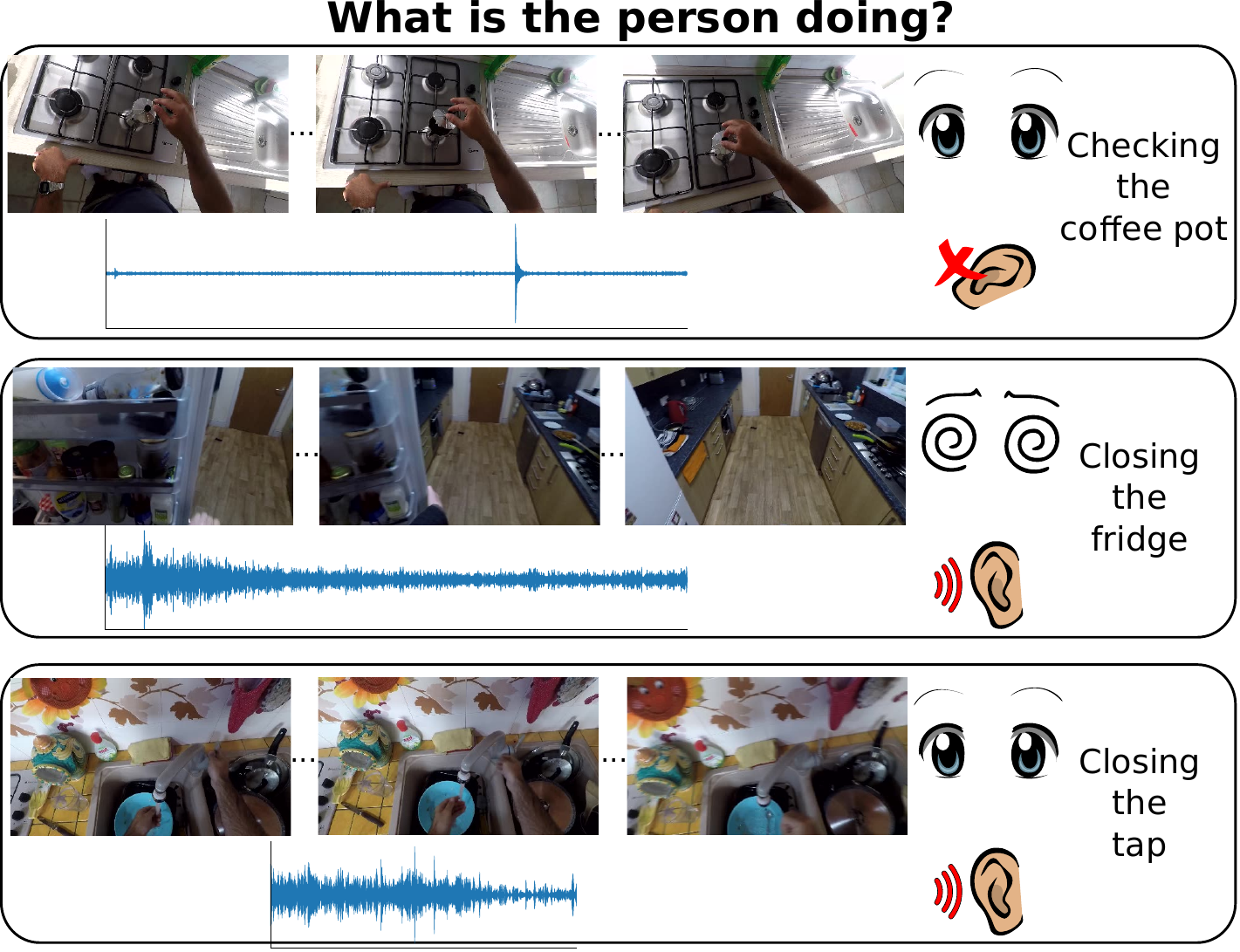}
\caption[]{Audio and vision are complementary sources of information for recognizing egocentric object interactions. A limited number of interactions do not have an associated audio signal (top), but in most cases, auditory sources provide valuable information in situations such as the occlusion of the hands and objects (middle), and in some others they just strengthen the visual information (bottom).}
\label{fig:seeingAndHearing}
\end{center}
\end{figure}

In this work, we focus on a particular case of activity recognition, that is the recognition of activities involving object manipulations. More specifically, the goal is to identify actions of the type \textit{verb}$+$\textit{noun}, i.e. \textit{pouring}$+$\textit{jam} performed in a kitchen environment. Recently, promising results on this task based solely on audio features have being shown in \cite{cartas2019much}. Motivated by this work, we proposed a framework to integrate visual and audio features. Fig.\;\ref{fig:seeingAndHearing} shows how audio cues are crucial to identify the activity being performed specially when visual information is ambiguous from an egocentric perspective due to self-occlusions. Our proposal aims at exploiting the complementarity of visual and audio features to obtain robust multimodal representations. 

\begin{figure*}[!t]
\begin{center}
\includegraphics[width=0.8\textwidth]{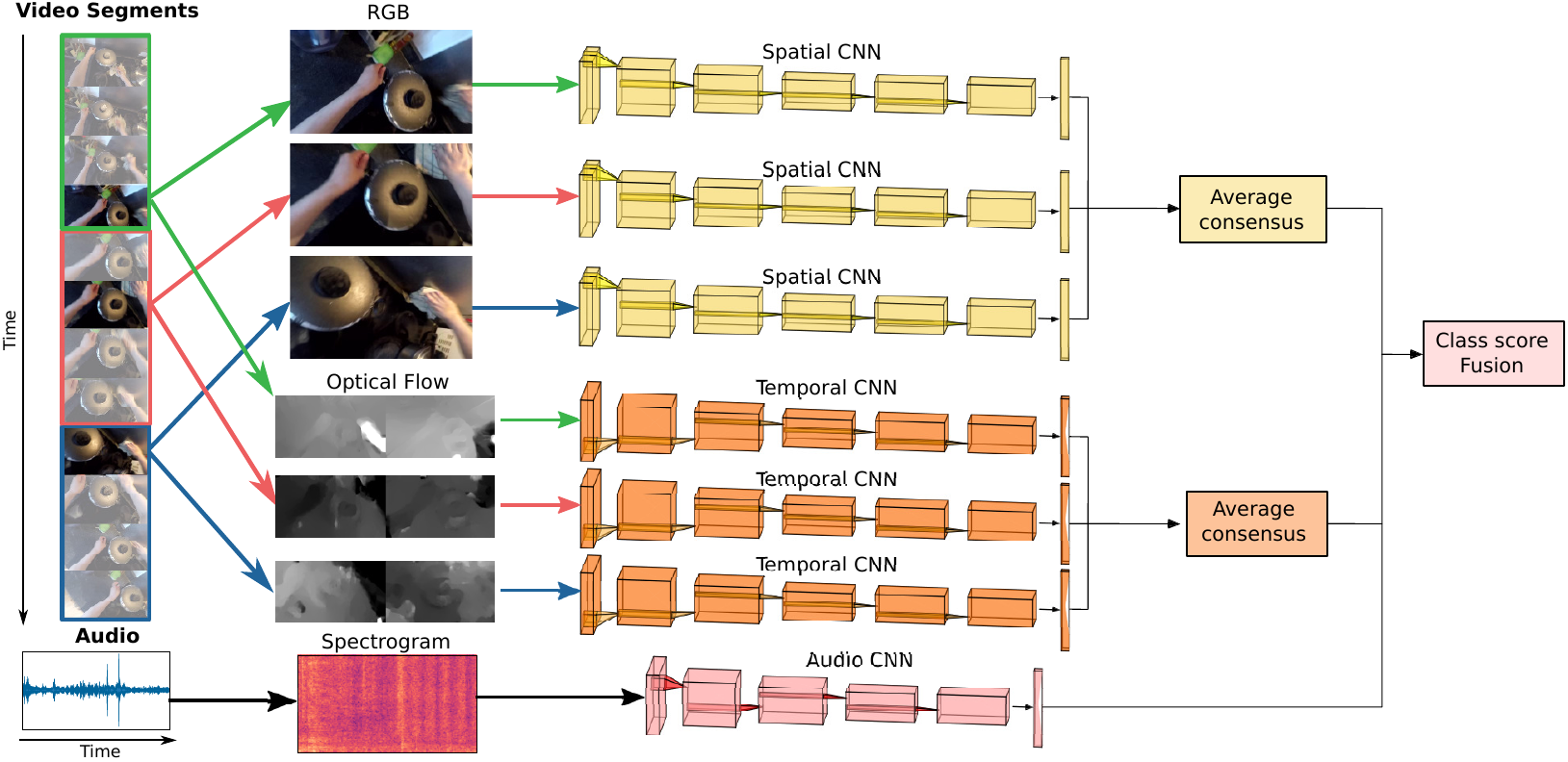}
\caption[]{Pipeline of our proposed approach. A video is divided into $K=3$ time segments shown in \textcolor{PipelineGreen}{\textbf{green}}, \textcolor{PipelineRed}{\textbf{red}}, and \textcolor{PipelineBlue}{\textbf{blue}} colors. Then, RGB and optical flow frames are sparsely sampled from each time segment to be processed in their respective spatial and temporal streams. At the end of each stream, the average consensus of the softmax scores is computed. A spectrogram is calculated from the raw audio signal and processed in its audio stream. The class scores of each stream are joined together using late fusion.}
\label{fig:pipeline}
\end{center}
\end{figure*}

In particular, our contributions can be summarized as follows: \one We provide an extensive evaluation and comparison with published methods of the proposed multimodal architecture on the EPIC-Kitchens dataset \cite{Damen2018EPICKITCHENS} \two In addition to action performance, we provide for the first time a detailed results on the \textit{object} and \textit{verb} components. The rest of the paper is organized as follows. The next section discusses related work. Section \ref{sec:approach} introduces the proposed approach, while section \ref{sec:setup} and section \ref{sec:results} detail the experimental setup and discuss the results, respectively. Finally, section \ref{sec:fine} concludes the paper by summarizing the main findings. 

\section{Related Work}
\label{sec:SoA}

\paragraph{Activity recognition}
The literature on activity recognition is vast and spans several decades \cite{herath2017going, poppe2010survey}. In the following, we will limit the discussion to egocentric object interactions recognition methods. Early approaches aimed at modelling spatio-temporal features through probabilistic models \cite{fathi2011understanding,fathi2012learning}, and temporal or spatio-temporal pyramids \cite{mccandless2013object,pirsiavash2012detecting}. Later on, multi-stream neural network based approaches were proposed \cite{Ma_2016_CVPR,Singh_2016_CVPR}. Typically, each stream treats a different modality (motion, RGB) or models different cues (hands, objects, etc). Attention-based mechanisms were proposed in \cite{sudhakaran2019lsta,sudhakaran2018}. The key idea underlying these approaches is learning to attend regions containing objects correlated with the activity under consideration. Recently, Baradel et al. \cite{Baradel_2018_ECCV} build on the output of an object detector to perform a spatio-temporal reasoning about the action being performed. 

Two relevant works have been focused on acoustic scene and activity classification from data collected in the wild \cite{Hwang2012, LiangAcousticADL2019}. Hwang and Lee \cite{Hwang2012} clustered thirteen acoustic classes based on Mel-frequency cepstrum coefficient (MFCC) features from data gathered using mobile phones. They used as a classification method a $k$-nearest neighborhood on hand-crafted histograms from the same spectral features. A closer work to ours is presented in \cite{LiangAcousticADL2019}. They proposed a model to classify 15 different one-minute home activities of daily living. A shallow network was trained using the encodings generated by pre-trained VGG-11 network \cite{HersheyAudioSetCNNs2017} and an oversampling strategy using a large audio set \cite{GemmekeAudioSet2017}.

\paragraph{Multisensory integration}
Most multimodal approaches for activity recognition through wearable sensors build on the combination of the several inertial features captured by accelerometers, gyroscopes and magnetic sensors \cite{ordonez2016deep}. Inertial cues are often integrated with visual information \cite{yu2019hierarchical}. Audiovisual integration has been extensively used in the context of smart room scene analysis \cite{shivappa2009hierarchical} and event detection in surveillance videos \cite{cristani2007audio}. More recently, the integration of audio and visual features has been successfully used for detecting human-to-human interactions \cite{bano2018multimodal}, for apparent personality trait recognition \cite{guccluturk2016deep}, for video description generation \cite{jin2016video}, and for multispeaker tracking \cite{qian2019multi}. Although several works that combine audiovisual sources have been reported in the context first-person action recognition challenges \cite{HeilbronActivityRecognitionChallenge2016,GhanemActivityRecognitionChallenge2018,GhanemActivityRecognitionChallenge2017}, they provide few details about their models. In \cite{LongAttentionClustersCVPR2018,LongMultimodalKeyless2018} are proposed attention mechanisms for action recognition using audio as a modality branch. However, the use of audio-visual cues for object interaction recognition is still very limited and previous works only reported results on the full interaction (\textit{action}) and not its components (\textit{verb} and \textit{noun}).

The three main model-agnostic approaches largely used in the literature for combining the audio-visual features are early, late, and hybrid fusion \cite{FusionMethods2019,KarpathyFusionMethodsVideo2014}. In the first approach, a model learns from the multimodal features after joining them \cite{jin2016video,PieropanAudioVisualCues2014,qian2019multi}. In the second approach, different unimodal predictors are trained and the final decision is made by combining their output \cite{guccluturk2016deep}. The last approach combines the output from early and late fusion predictors \cite{cristani2007audio,shivappa2009hierarchical}.

\section{Proposed approach}
\label{sec:approach}
In this section we describe our multimodal approach for action recognition\footnote{Code at: \url{http://github.com/gorayni/seeing_and_hearing}}, that is summarized in Fig. \ref{fig:pipeline}. We first present our vision-based recognition approach that uses Temporal Segments Network (TSN)\cite{WangTSN2016} for the spatial (RGB) and temporal (optical flow) visual modalities. Then, we present our audio-based recognition approach that uses two different convolutional neural networks (CNNs): VGG-11\cite{Simonyan14c} and a custom network based on \cite{Sainath2015}. Finally, we detail our fusion strategy to integrate the different modalities.

\paragraph{Vision}
The visual spatial and temporal input modalities are RGB and optical flow frames calculated using \cite{ZachOpticalFlow2007}. Each visual modality was trained as a TSN stream. On the TSN model, the frames of a video are grouped into $K$ sequential segments of equal size. Similarly to \cite{WangTSN2016}, we decided to set $K=3$ as originally presented. Simultaneously from each segment, a frame is sparsely sampled and processed by a CNN. In our case, we used as backbone networks a ResNet-18 and ResNet-50 \cite{He_2016_CVPR} in our experiments. Then, a consensus of the scores from each processed frame is done. We used as a consensus function the average of the softmax scores. This model is an extension of \cite{simonyan2014two}, but it learns long-range temporal structure of the action in the video.

\paragraph{Audio}
The audio modality uses as input the spectrogram of the raw audio signal from the video. The spectrogram is calculated as follows. First, when the video has multiple audio channels, we join them by obtaining their mean. Then, we compute the short-time Fourier transform (STFT) from this signal using a sampling frequency of $16$ KHz. The STFT uses a Hamming window of length equal to 30 ms with $50\%$ time overlapping. The signal spectrogram is calculated as the logarithm value of the squared magnitude of its STFT. The final step consists in normalizing all the input spectrograms. The spectrogram has a resulting dimension size for the frequency of 331. As in \cite{cartas2019much}, we only consider the first four seconds of the audio spectrogram. When it has less than four seconds duration then a zero padding is applied. This constraint results in a time dimension size of 248 for the input spectrogram.

A single spectrogram covers a larger time window than the visual input frames. Therefore, our model only needs one CNN to process the audio modality. Nonetheless, for longer video durations a long short-term memory (LSTM) could be added as in \cite{SainathLSTMAudio2015}. Our backbone audio CNN models are a VGG-11\cite{Simonyan14c} network and a proposed smaller CNN based on \cite{SainathSmallAudioCNN2015}. We call the latter traditional dilated network and show its architecture on Table \ref{tab:traditionalDilatedNetwork}. This network was adapted to spectrograms with bigger sizes by using dilation convolutions \cite{YuKoltun2016}.

\begin{table}
\centering
\resizebox{0.75\columnwidth}{!}{%
\begin{tabular}{|l|c|c|c|c|}
\hline
Layer type & Output size& \#Filters & Kernel size & Dilation \\\hline
Conv2D  & 331$\times$248  &  64 & 11$\times$7 & 9$\times$4   \\ \hline
\multicolumn{5}{|c|}{max-pool} \\ \hline
Conv2D  & 166$\times$124  & 64 & 6$\times$4 & 9$\times$4 \\ \hline
Conv2D  & 166$\times$124  & 32 & 6$\times$4 & 9$\times$4 \\ \hline
Conv2D  & 166$\times$124  & 16 & 6$\times$4 & 9$\times$4 \\ \hline
\multicolumn{5}{|c|}{max-pool} \\ \hline
Dense & 256 & - & - & -  \\ \hline
Dense & 256 & - & - & -  \\ \hline
\end{tabular}
}
\caption{Architecture of the traditional dilated network for audio classification.}
\label{tab:traditionalDilatedNetwork}
\end{table}

\paragraph{Audio-visual fusion}

In our experiments we used two late fusion methods. The first method is the weighted sum of the class scores from each stream. The second method uses a network with two fully connected (FC) layers. Its input vector is calculated by concatenating the outputs of the penultimate FC layers from each stream. During training, the weights of each modality CNN stream are kept frozen.

\begin{table*}[!t]
\begin{center}
\resizebox{0.9\textwidth}{!}{%
\begin{tabular}{lccl|ccc|ccc|ccc|ccc}
& & &%
&\multicolumn{3}{c|}{\textbf{Top-1 Accuracy}}%
&\multicolumn{3}{c|}{\textbf{Top-5 Accuracy}}%
&\multicolumn{3}{c|}{\textbf{Avg Class Precision}}%
&\multicolumn{3}{c}{\textbf{Avg Class Recall}}\\\cline{5-16}
& & &%
&VERB &NOUN &ACTION%
&VERB &NOUN &ACTION%
&VERB &NOUN &ACTION%
&VERB &NOUN &ACTION\\ \cline{4-16}
& & & Chance/Random & 11.38 & 01.58 & 00.43 & 47.58 & 07.74 & 02.12 & 01.00 & 00.34 & 00.06 & 01.00 & 00.34 & 00.07 \\ \vspace{5mm}
& & & Largest class & 20.19 & 04.11 & 02.10 & 66.93 & 18.38 & 07.54 & 00.21 & 00.01 & 00.00 & 01.05 & 00.36 & 00.07 \\ \cline{2-16}
\multirow{14}{*}{\rotatebox{90}{\textbf{ACTION}}}%
& \textbf{Audio}%
& & VGG-11%
& \textcolor{TableGray}{29.44} & \textcolor{TableGray}{05.99} & 05.84%
& \textcolor{TableGray}{70.49} & \textcolor{TableGray}{16.90} & 17.41%
& \textcolor{TableGray}{00.92} & \textcolor{TableGray}{00.51} & 00.20%
& \textcolor{TableGray}{01.86} & \textcolor{TableGray}{00.79} & 00.45\\
& & & Traditional Dilated%
& \textcolor{TableGray}{32.16} & \textcolor{TableGray}{06.86} & 10.77%
& \textcolor{TableGray}{72.22} & \textcolor{TableGray}{22.34} & 25.86%
& \textcolor{TableGray}{01.64} & \textcolor{TableGray}{00.84} & 03.04%
& \textcolor{TableGray}{02.31} & \textcolor{TableGray}{01.90} & 04.58\\\cline{2-16}
& & & TSN Flow%
& \textcolor{TableGray}{36.70} & \textcolor{TableGray}{14.82} & 22.54%
& \textcolor{TableGray}{75.98} & \textcolor{TableGray}{37.25} & 41.52%
& \textcolor{TableGray}{02.94} & \textcolor{TableGray}{03.46} & 05.98%
& \textcolor{TableGray}{03.89} & \textcolor{TableGray}{03.06} & 07.04 \\
& \textbf{Vision}%
& & TSN RGB%
& \textcolor{TableGray}{34.06} & \textcolor{TableGray}{31.01} & 32.80%
& \textcolor{TableGray}{76.71} & \textcolor{TableGray}{63.90} & 59.06%
& \textcolor{TableGray}{04.17} & \textcolor{TableGray}{13.45} & 15.37%
& \textcolor{TableGray}{04.83} & \textcolor{TableGray}{11.16} & 19.62 \\
& & & RGB+Flow%
& \textcolor{TableGray}{37.12} & \textcolor{TableGray}{30.15} & 36.19%
& \textcolor{TableGray}{79.56} & \textcolor{TableGray}{62.02} & 59.92%
& \textcolor{TableGray}{04.96} & \textcolor{TableGray}{11.94} & 15.14%
& \textcolor{TableGray}{04.19} & \textcolor{TableGray}{09.21} & 17.84\\ \cline{2-16}
& &%
& Flow+Audio%
& \textcolor{TableGray}{38.04} & \textcolor{TableGray}{13.03} & 25.35%
& \textcolor{TableGray}{77.42} & \textcolor{TableGray}{35.81} & 45.06%
& \textcolor{TableGray}{03.89} & \textcolor{TableGray}{04.52} & 09.25%
& \textcolor{TableGray}{03.59} & \textcolor{TableGray}{03.21} & 09.32\\
& \multirow{4}{*}{\textbf{Multimodal}}%
& & RGB+Audio%
& \textcolor{TableGray}{37.65} & \textcolor{TableGray}{27.03} & 34.93%
& \textcolor{TableGray}{79.52} & \textcolor{TableGray}{59.41} & 60.69%
& \textcolor{TableGray}{04.03} & \textcolor{TableGray}{13.88} & 16.08%
& \textcolor{TableGray}{03.66} & \textcolor{TableGray}{09.49} & 18.50\\
& & & RGB+Flow+Audio%
& \textcolor{TableGray}{39.95} & \textcolor{TableGray}{27.45} & 36.78%
& \textcolor{TableGray}{80.47} & \textcolor{TableGray}{59.17} & 60.38%
& \textcolor{TableGray}{03.30} & \textcolor{TableGray}{11.03} & 15.51%
& \textcolor{TableGray}{04.02} & \textcolor{TableGray}{08.26} & 16.96\\ \cline{3-16}
& &\multirow{3}{*}{\rotatebox{90}{\footnotesize Weighted}}%
& Flow+Audio%
& \textcolor{TableGray}{39.66} & \textcolor{TableGray}{14.44} & 26.03%
& \textcolor{TableGray}{77.53} & \textcolor{TableGray}{39.15} & 46.16%
& \textcolor{TableGray}{03.60} & \textcolor{TableGray}{04.78} & 09.30%
& \textcolor{TableGray}{03.97} & \textcolor{TableGray}{03.39} & 09.29\\
& & & RGB+Audio%
& \textcolor{TableGray}{38.80} & \textcolor{TableGray}{30.15} & 35.50%
& \textcolor{TableGray}{80.31} & \textcolor{TableGray}{63.44} & 61.71%
& \textcolor{TableGray}{04.22} & \textcolor{TableGray}{14.97} & 16.71%
& \textcolor{TableGray}{04.20} & \textcolor{TableGray}{10.91} & 19.63\\
& & & RGB+Flow+Audio%
& \textcolor{TableGray}{40.06} & \textcolor{TableGray}{29.68} & 36.92%
& \textcolor{TableGray}{\textbf{80.82}} & \textcolor{TableGray}{61.69} & 61.14%
& \textcolor{TableGray}{03.16} & \textcolor{TableGray}{12.47} & 15.56%
& \textcolor{TableGray}{04.12} & \textcolor{TableGray}{09.47} & 17.76\\\cline{3-16}
& &\multirow{3}{*}{\rotatebox{90}{\footnotesize FC}}%
& Flow+Audio%
& \textcolor{TableGray}{41.58} & \textcolor{TableGray}{21.26} & 27.34%
& \textcolor{TableGray}{79.63} & \textcolor{TableGray}{48.33} & 47.84%
& \textcolor{TableGray}{05.81} & \textcolor{TableGray}{07.16} & 12.63%
& \textcolor{TableGray}{05.08} & \textcolor{TableGray}{05.69} & 14.01\\
& & & RGB+Audio%
& \textcolor{TableGray}{40.85} & \textcolor{TableGray}{36.39} & 35.94%
& \textcolor{TableGray}{76.84} & \textcolor{TableGray}{70.05} & 61.31%
& \textcolor{TableGray}{\textbf{09.03}} & \textcolor{TableGray}{17.13} & 16.40%
& \textcolor{TableGray}{07.11} & \textcolor{TableGray}{12.88} & 19.57\\\vspace{5mm}
& & & RGB+Flow+Audio%
& \textcolor{TableGray}{\textbf{42.56}} & \textcolor{TableGray}{\textbf{36.81}} & \textbf{40.15}%
& \textcolor{TableGray}{77.06} & \textcolor{TableGray}{\textbf{70.38}} & \textbf{64.19}%
& \textcolor{TableGray}{08.48} & \textcolor{TableGray}{\textbf{18.08}} & \textbf{19.21}%
& \textcolor{TableGray}{\textbf{07.55}} & \textcolor{TableGray}{\textbf{12.93}} & \textbf{22.68}\\ \cline{2-16}
\multirow{14}{*}{\rotatebox{90}{\textbf{VERB+NOUN}}}%
& \textbf{Audio}%
& & VGG-11%
& 34.48 & 09.51 & \textcolor{TableGray}{03.56}%
& 74.50 & 26.63 & \textcolor{TableGray}{12.17}%
& 05.26 & 01.32 & \textcolor{TableGray}{00.28}%
& 04.04 & 01.32 & \textcolor{TableGray}{01.72} \\
& & & Traditional Dilated%
& 34.82 & 15.44 & \textcolor{TableGray}{06.26}%
& 74.72 & 36.96 & \textcolor{TableGray}{17.83}%
& 04.53 & 05.77 & \textcolor{TableGray}{01.12}%
& 03.88 & 04.95 & \textcolor{TableGray}{01.39} \\ \cline{2-16}
& & & TSN Flow%
& 49.08 & 22.72 & \textcolor{TableGray}{13.54}%
& 81.60 & 46.32 & \textcolor{TableGray}{30.77}%
& 10.80 & 08.81 & \textcolor{TableGray}{02.53}%
& 07.12 & 04.97 & \textcolor{TableGray}{02.23} \\
& \textbf{Vision}%
& & TSN RGB%
& 50.65 & 54.01 & \textcolor{TableGray}{32.51}%
& 88.63 & 80.87 & \textcolor{TableGray}{59.72}%
& 25.96 & 38.83 & \textcolor{TableGray}{16.23}%
& \textbf{19.36} & \textbf{34.43} & \textcolor{TableGray}{18.94} \\
& & & RGB+Flow%
& 55.47 & 52.82 & \textcolor{TableGray}{32.76}%
& 88.48 & 78.01 & \textcolor{TableGray}{58.13}%
& 28.94 & 39.82 & \textcolor{TableGray}{13.53}%
& 14.25 & 27.81 & \textcolor{TableGray}{14.22} \\ \cline{2-16}
& &%
& Flow+Audio%
& 50.06 & 26.26 & \textcolor{TableGray}{15.13}%
& 81.02 & 51.45 & \textcolor{TableGray}{33.49}%
& 11.72 & 11.32 & \textcolor{TableGray}{02.99}%
& 06.61 & 06.42 & \textcolor{TableGray}{02.42} \\
& \multirow{4}{*}{\textbf{Multimodal}}%
& & RGB+Audio%
& 53.51 & 53.11 & \textcolor{TableGray}{32.21}%
& 87.35 & 79.63 & \textcolor{TableGray}{57.82}%
& 26.57 & 38.89 & \textcolor{TableGray}{13.67}%
& 13.07 & 28.98 & \textcolor{TableGray}{14.94} \\
& & & RGB+Flow+Audio%
& 56.27 & 51.09 & \textcolor{TableGray}{32.27}%
& 87.24 & 77.15 & \textcolor{TableGray}{55.96}%
& 25.06 & 37.17 & \textcolor{TableGray}{11.81}%
& 11.34 & 24.28 & \textcolor{TableGray}{12.05} \\ \cline{3-16}
& &\multirow{3}{*}{\rotatebox{90}{\footnotesize Weighted}}%
& Flow+Audio%
& 51.40 & 26.39 & \textcolor{TableGray}{15.62}%
& 81.57 & 51.54 & \textcolor{TableGray}{34.24}%
& 11.86 & 12.57 & \textcolor{TableGray}{03.37}%
& 06.98 & 06.48 & \textcolor{TableGray}{02.72} \\
& & & RGB+Audio%
& 54.24 & 54.90 & \textcolor{TableGray}{33.84}%
& 88.19 & 80.89 & \textcolor{TableGray}{59.72}%
& \textbf{31.21} & 38.96 & \textcolor{TableGray}{15.03}%
& 15.07 & 31.74 & \textcolor{TableGray}{16.73} \\
& & & RGB+Flow+Audio%
& 56.65 & 53.90 & \textcolor{TableGray}{33.86}%
& 87.70 & 79.47 & \textcolor{TableGray}{58.37}%
& 25.75 & \textbf{40.87} & \textcolor{TableGray}{13.36}%
& 12.58 & 28.17 & \textcolor{TableGray}{14.02} \\ \cline{3-16}
& &\multirow{3}{*}{\rotatebox{90}{\footnotesize FC}}%
& Flow+Audio%
& 52.00 & 33.02 & \textcolor{TableGray}{20.22}%
& 82.24 & 58.50 & \textcolor{TableGray}{39.37}%
& 09.72 & 18.58 & \textcolor{TableGray}{05.80}%
& 08.11 & 16.63 & \textcolor{TableGray}{06.58} \\
& & & RGB+Audio%
& 55.41 & 55.08 & \textcolor{TableGray}{35.21}%
& 87.15 & 79.78 & \textcolor{TableGray}{60.27}%
& 20.49 & 38.47 & \textcolor{TableGray}{15.35}%
& 14.28 & 33.55 & \textcolor{TableGray}{17.64}\\
& & & RGB+Flow+Audio%
& \textbf{60.21} & \textbf{56.14} & \textcolor{TableGray}{\textbf{38.55}}%
& \textbf{89.07} & \textbf{80.96} & \textcolor{TableGray}{\textbf{62.97}}%
& 27.05 & 39.89 & \textcolor{TableGray}{\textbf{17.14}}%
& 19.09 & 33.85 & \textcolor{TableGray}{\textbf{19.28}} \\
\end{tabular}
}
\caption{Classification performance for the equally stratified \textit{action} data split. The \textit{action} part show the results of training one classifier on the action labels, whereas the \textit{verb+noun} part show the results of independently training two classifiers over the verb and noun labels. The scores in \textcolor{TableGray}{\textbf{gray}} color were calculated based on the respective \textit{action} or \textit{verb+noun} classifiers.}
\label{tab:actionRecognition}
\end{center}
\vspace{-5mm}
\end{table*}

\begin{figure*}[h!]
\centering
\includegraphics[width=1\textwidth]{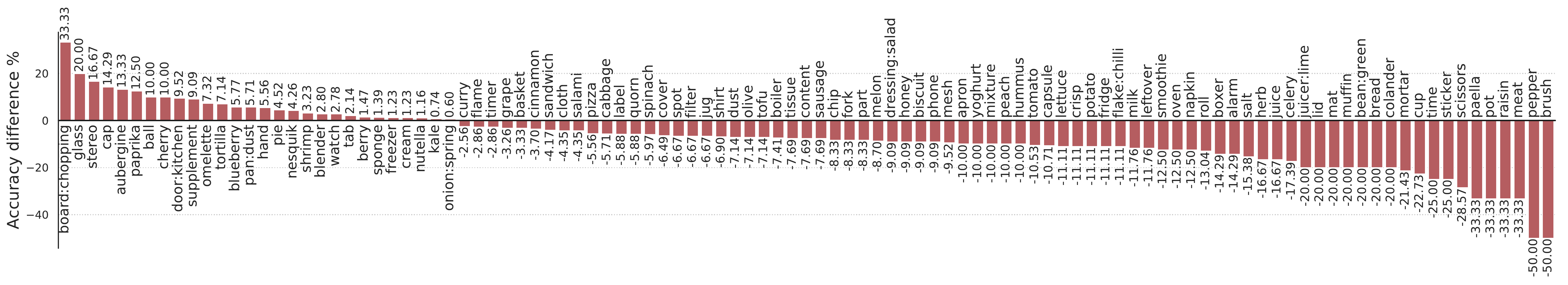}
\caption{Accuracy difference of the \textit{noun} split for the unweighted test predictions that changed with respect to RGB+Flow and RGB+Flow+Audio.}
\label{fig:nounAccDiff}
\end{figure*}

\section{Experimental Setup}
\label{sec:setup}
The main objective of our experiments was to measure the performance of our proposed multimodal approach on an egocentric object interaction recognition task from videos. More specifically, the task consists on predicting what a person is doing (\textit{verb}) using a specific object (\textit{noun}). Both classifications can be trained and evaluated separately or combined as a single \textit{action} classification. Therefore, our secondary objective was to determine the contribution of audio and visual information on each type of classification (noun, verb, action). Contrary to previous works \cite{HeilbronActivityRecognitionChallenge2016,GhanemActivityRecognitionChallenge2018,GhanemActivityRecognitionChallenge2017}, we did not make any assumption on which classification type the audio source would perform better. We provide further details of our experimental setup in the following subsections. In section \ref{sec:dataset}, we describe the used dataset and its data partition. The specific evaluation metrics and baselines are presented in section \ref{sec:evaluationMetrics}. Finally, the implementation details of our model are detailed in section \ref{sec:implementation}.

\subsection{Dataset}
\label{sec:dataset}

We carried out our experiments on the EPIC Kitchens dataset \cite{Damen2018EPICKITCHENS}. Each video segment in the dataset shows a participant doing one specific cooking related action in a kitchen environment. Some examples of their labels are ``cut potato'' or ``wash cup''. The EPIC Kitchens dataset includes 432 videos recorded from a first-person perspective by 32 participants in their own kitchens while cooking/preparing something. Each video was divided into segments in which the person is doing one specific \textit{action} (a \textit{verb} plus a \textit{noun}). The total number of verbs and nouns categories in dataset is 125 and 352, correspondingly.

For comparison purposes, we considered two data partitions derived from the labeled data of the EPIC Kitchen Challenge:

\paragraph{Home made partition} In this partition, all the participants were considered for the training, validation, and test splits. The data proportions for the validation and test splits were 10\% and 15\%, accordingly. Since the resulting distribution of action classes is highly unbalanced, the data split was done as follows. At least one sample of each action category was put in the training split. If the category had at least two samples, one of them went to the test split. We also report the results obtained on the EPIC Kitchens Challenge board from the models trained on this partition.

\begin{figure*}[t!]
\centering
\includegraphics[width=0.6\textwidth]{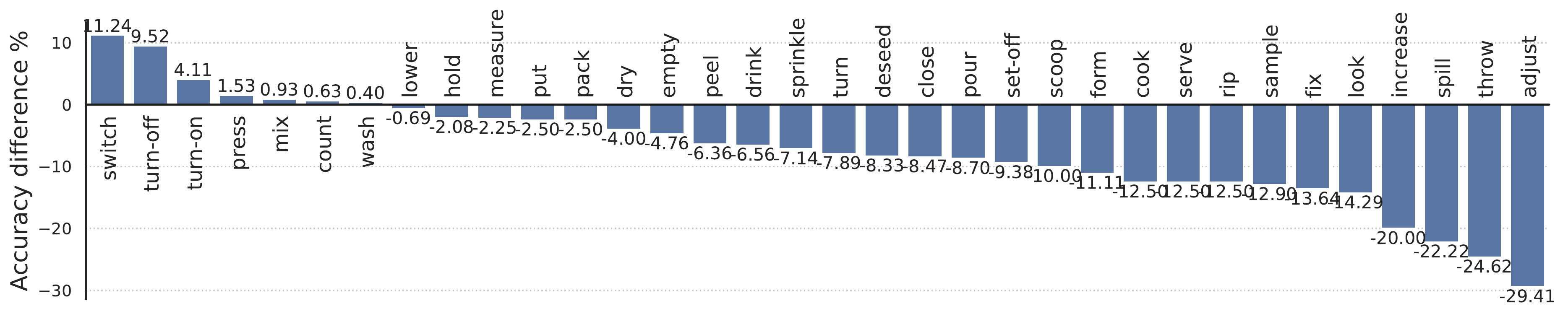}
\caption{Accuracy difference of the \textit{verb} split for the unweighted test predictions that changed with respect to RGB+Flow and RGB+Flow+Audio.}
\label{fig:verbAccDiff}
\end{figure*}

\begin{figure}[!h]
\begin{center}
\includegraphics[width=1\columnwidth]{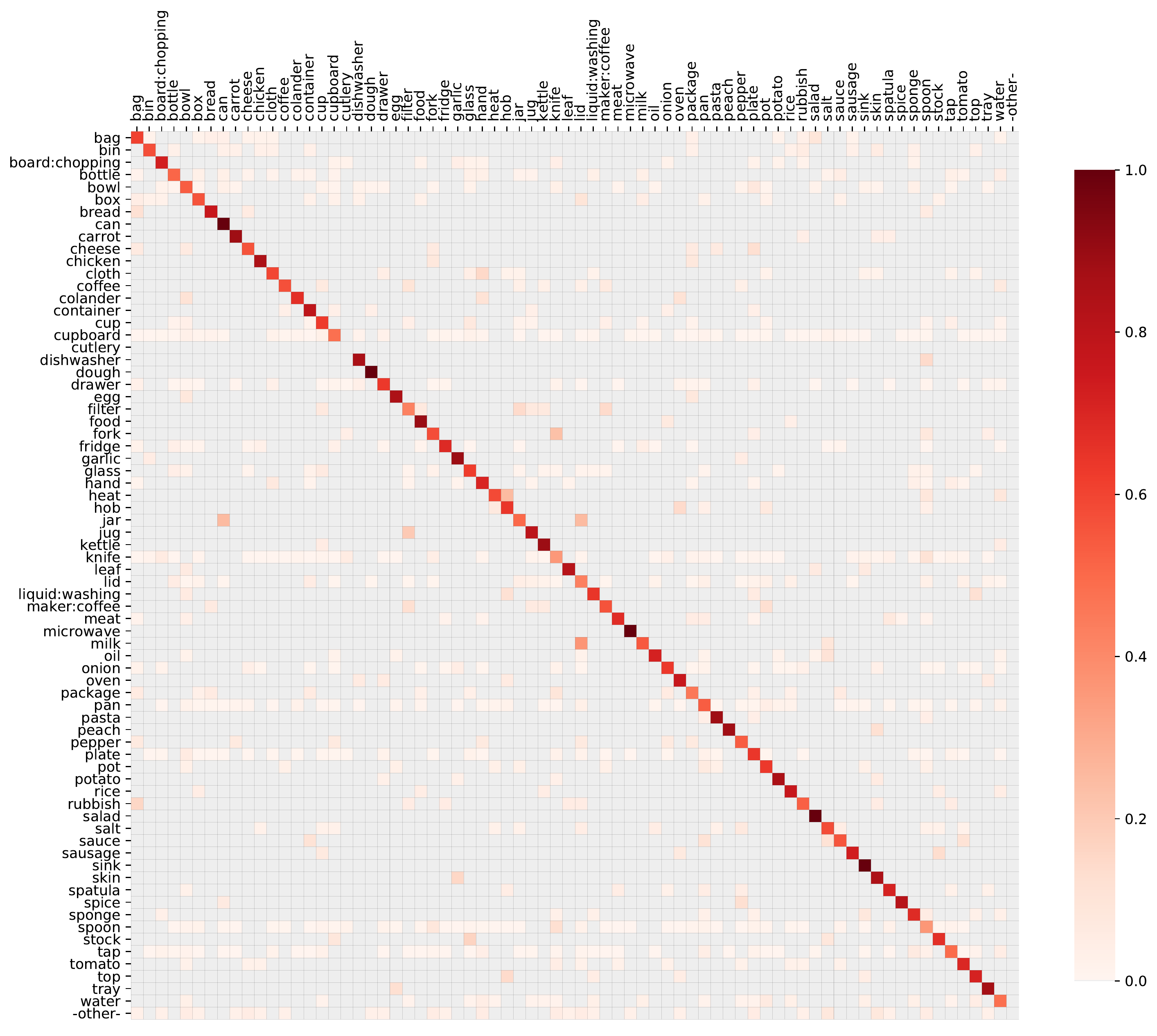}
\caption[]{Normalized noun confusion matrix with an accuracy of $61.94\%$. Only the categories with more than 100 samples are shown.}
\label{fig:nounConfusionMatrix}
\end{center}
\end{figure}

\paragraph{Unseen verb partition} The second data partition was the one proposed \cite{Baradel_2018_ECCV}. This partition only used the \textit{verb} classes from the labeled data. Moreover, the training and test splits are on the participants 01-25 and 26-31, respectively. Therefore, the test set is only composed of unseen kitchens. In order to train our methods, we created a randomly stratified validation split with 10\% of data from the training split.

\subsection{Evaluation metrics}
\label{sec:evaluationMetrics}
Following \cite{Damen2018EPICKITCHENS}, we measured the classification performance using aggregate and per-class metrics. As aggregate metrics for measuring the classification performance we used the top-1 and top-5 accuracy, whereas as per-class metrics we used precision and recall. Moreover, the per-class classification improvement was measured by calculating the accuracy difference between the visual (RGB+Flow) and the audiovisual (RGG+Flow+Audio) sources. We also computed two baselines using the largest classes and random classifiers for each experiment. The latter baseline was approximated by sampling a multinomial distribution.

\subsection{Implementation}
\label{sec:implementation}

We first trained all modality streams separately on each training split for \textit{verb}, \textit{noun}, and \textit{action}. Subsequently, we trained their late fusion on different combinations of audio and vision streams. On our experiments, we searched for the best learning rates while performing early stopping using the validation split. The following paragraphs provide more training details for each part of the model.

\paragraph{Audio} We only trained our audio network on the spectrogram of the first four seconds of each video segment. As stated in \cite{cartas2019much}, setting a time threshold of 4 seconds allows to completely cover $80.697\%$ of all video segments using a single time window. For all our experiments we used the stochastic gradient descent (SGD) optimization algorithm to train both networks from scratch. We used a momentum and a batch size equal to 0.9 and 6, correspondingly. The learning rates for VGG-11\cite{Simonyan14c} on \textit{verb}, \textit{noun}, and \textit{action} classification were $5\times10^{-6}$, $2.5\times10^{-6}$, and $1.75\times10^{-6}$, respectively. The learning rates for the traditional dilated network on \textit{verb}, \textit{noun}, and \textit{action} classification were $4.5\times10^{-4}$, $7.5\times10^{-5}$, and $1\times10^{-4}$, accordingly. It was trained using a learning rate equal $1\times10^{-5}$ and a batch size of 22 during 65 epochs. The difference between the different data splits was the number of training epochs. The training times for the VGG-11 and the Traditional Dilated network were around fifteen and eleven hours on a Nvidia GeForce GTX 980, respectively.

\begin{table*}[t]
\centering
\resizebox{1.7\columnwidth}{!}{%
\begin{tabular}{lcl|c|c|c|c}
\cline{3-7}
& & \multicolumn{1}{c|}{\textbf{Method}} & \textbf{Top-1 Accuracy} & \textbf{Top-5 Accuracy} & \textbf{Avg. Class Precision} & \textbf{Avg. Class Recall} \\ \hline
& & Chance/Random & 11.75 & 48.87 & 00.99 & 00.98 \\ 
& & Largest class & 21.27 & 69.44 & 00.31 & 01.41 \\ \hline
\multirow{2}{*}{\rotatebox{90}{\small\textbf{Audio}}}%
& & Traditional Dilated & 30.51 & 74.19 & 04.71 & 03.60 \\[0.2ex]
& & VGG-11 & 33.27 & 74.13 & 05.72 & 04.08 \\ \hline
\multirow{6}{*}{\rotatebox{90}{\small\textbf{Vision}}}%
& & ResNet-18 ~\cite{He_2016_CVPR}$\dagger$ & 32.05 & - & - & - \\
& & I3D ResNet-18~\cite{Carreira_2017_CVPR}$\dagger\;$ & 34.20 & - & - & -\\
& & TSN ResNet-18 RGB & 34.69 & 77.13 & 08.38 & 05.08 \\
& & ORN~\cite{Baradel_2018_ECCV}$\dagger\;$ & 40.89 & - & - & - \\
& & TSN ResNet-18 Flow & 44.48 & 77.88 & 08.15 & 06.18 \\
& & RGB+Flow & 43.36 & 78.77 & 09.98 & 05.58 \\ \hline
\multirow{6}{*}{\rotatebox{90}{\small\textbf{Multimodal}}}%
& & RGB+Audio VGG-11 & 41.09 & 80.10 & 08.82 & 04.76 \\
& & Flow+Audio VGG-11 & 45.75 & 80.34 & 08.41 & 05.57 \\
& & RGB+Flow+Audio VGG-11 & 45.86 & 80.72 & 09.33 & 05.25 \\\cline{2-7}
&\multirow{3}{*}{\rotatebox{90}{\footnotesize Weighted}}%
  & RGB+Audio VGG-11 & 40.83 & 79.42 & 09.45 & 04.85 \\
& & Flow+Audio VGG-11 & 43.74 & 79.80 & 09.68 & 05.31 \\
& & RGB+Flow+Audio VGG-11 & \textbf{46.07} & \textbf{80.76} & 09.34 & 05.30 \\\cline{2-7}
&\multirow{3}{*}{\rotatebox{90}{\footnotesize FC}}%
  & RGB+Audio VGG-11 & 42.08 & 79.44 & \textbf{12.51} & 06.55 \\
& & Flow+Audio VGG-11 & 44.50 & 78.68 & 08.17 & 06.86 \\
& & RGB+Flow+Audio VGG-11 & 45.92 & 80.33 & 11.01 & \textbf{07.31} \\
\hline
\end{tabular}
}
\caption{Classification performance results on the comparison \textit{verb} data split. The results marked with $\dagger$ were originally reported in \cite{Baradel_2018_ECCV}.}
\label{tab:actionRecognitionBaradel}
\end{table*}

\begin{figure*}[t]
\centering
\includegraphics[width=1\textwidth]{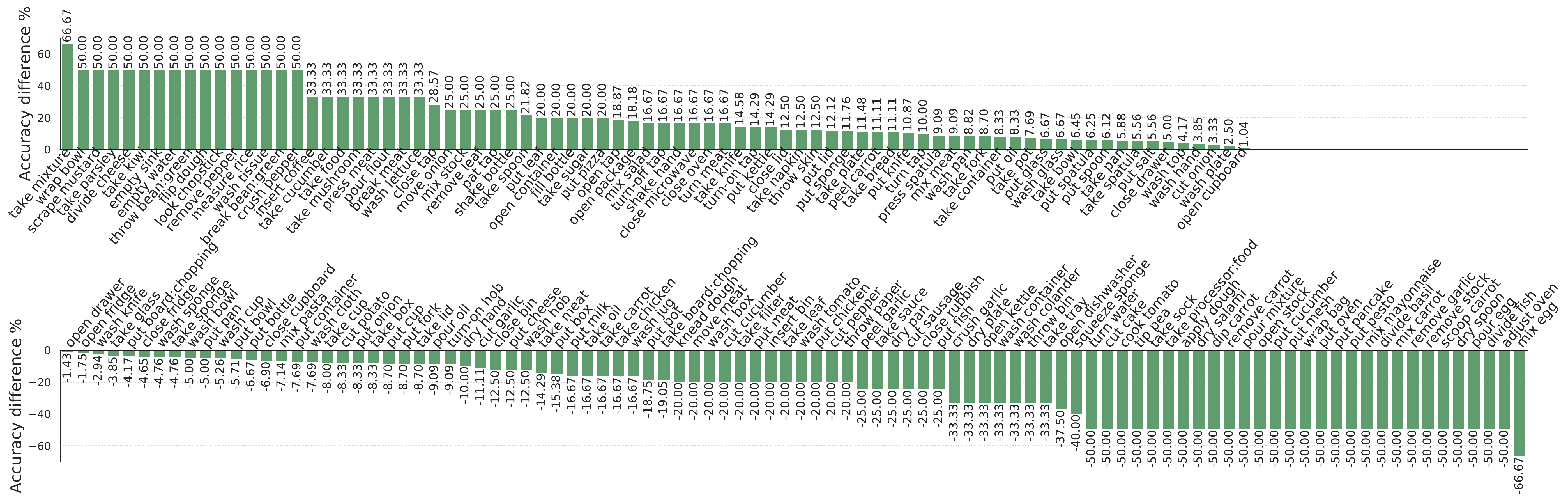}
\caption{Accuracy difference of the \textit{action} split for the unweighted test predictions that changed with respect to RGB+Flow and RGB+Flow+Audio.}
\label{fig:actionAccDiff}
\end{figure*}

\paragraph{Vision} We followed similar training specifications used in \cite{Damen2018EPICKITCHENS}, but considering the spatial and temporal CNNs as single networks rather than two joined streams. As previously stated, we used ResNet-18 and ResNet-50 as backbone CNNs. The former was used only in the data split presented in \cite{Baradel_2018_ECCV} for comparison purposes, while the latter was used in all other experiments. Each backbone CNN was used for both visual streams and were initialized using pre-trained weights on ImageNet \cite{DengImageNet2009}. Moreover, we trained both visual modalities between 40 and 80 epochs using the same learning rate of $1\times10^{-3}$ and decreasing it by a factor of 10 after epochs 20 and 40. The tests were done using 25 samples with 1 spatial cropping. The training of each modality and category took approximately twelve hours on a Nvidia Titan X GPU.

\paragraph{Audio-visual fusion} For the weighted sum of class scores, the weights were found using a grid search of values between 1 and 2. For the neural network method, depending of the backbone network used on each modality, the length of the input vector of the first FC layer was between 4,352 and 5,120. The second FC layer had and input vector length of 512. We used a momentum and a batch size equal to 0.9 and 6, correspondingly. The learning rates for the fusion of all modalities on \textit{verb}, \textit{noun}, and \textit{action} classification were $1\times10^{-4}$, $1\times10^{-3}$, and $3\times10^{-4}$, respectively.

\section{Results}
\label{sec:results}

\paragraph{Noun}
The performance results for the \textit{noun} classification on the home made partition are shown in the lower part of Table \ref{tab:actionRecognition}. The best accuracy score was achieved by the weighted multimodal combination that improved the visual baseline by 1.24\%. These results also indicate that the separated or combined unweighted fusion of the optical flow and audio decreases the top-1 and top-5 accuracy of the task. This effect can also be seen on the higher number of classes that decreased their accuracy on Fig. \ref{fig:nounAccDiff}. The most misclassified pair of objects are \textit{spoon}-\textit{knife}, \textit{spoon}-\textit{fork}, \textit{plate}-\textit{bowl}, \textit{tap}-\textit{sponge}, and \textit{knife}-\textit{fork}, as shown in Fig. \ref{fig:nounConfusionMatrix}.

\begin{figure}[h!]
\begin{center}
\includegraphics[width=0.8\columnwidth]{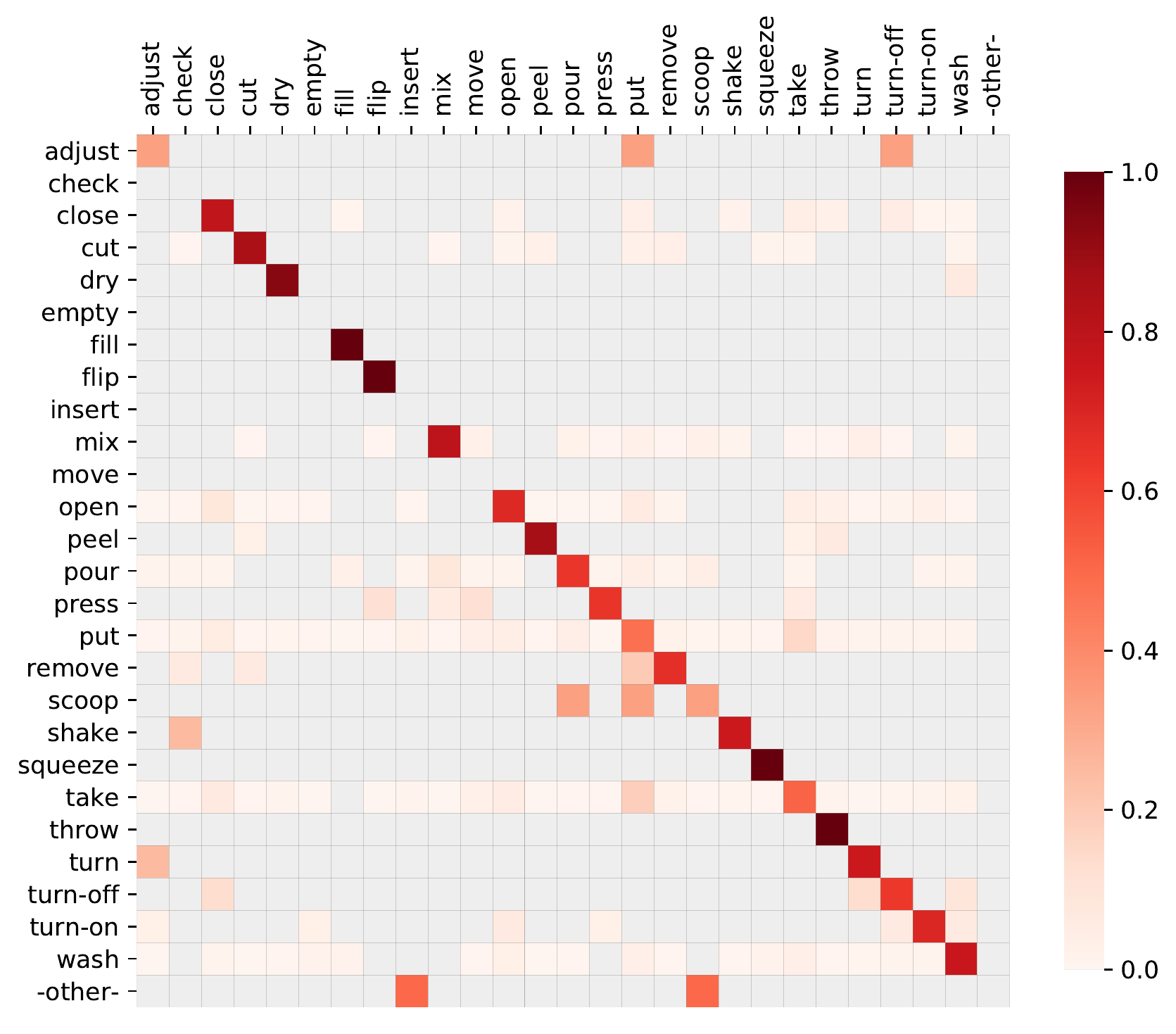}
\caption[]{Normalized verb confusion matrix with an accuracy of $62.43\%$. Only the categories with more than 100 samples are shown.}
\label{fig:verbConfusionMatrix}
\end{center}
\end{figure}

\paragraph{Verb}
The \textit{verb} classification results on the home made partition are also presented in the lower part of Table \ref{tab:actionRecognition}. Each visual method combination boosted their respective performance by adding the unweighted audio score. The multimodality combination is greater than the best visual method by 4.74\%. According to Fig. \ref{fig:verbAccDiff}, the multimodality helps to disambiguate \textit{turn-on} and \textit{turn-off} verbs, but fails on verbs that lack of sound like \textit{scoop} or \textit{adjust}. Fig. \ref{fig:verbConfusionMatrix} shows that the most misclassified pair of verbs are \textit{take}-\textit{put}, \textit{put}-\textit{open}, \textit{take}-\textit{close}, \textit{take}-\textit{open}, and \textit{put}-\textit{close}.

\begin{figure*}[!t]
\begin{center}
\includegraphics[width=0.81\textwidth]{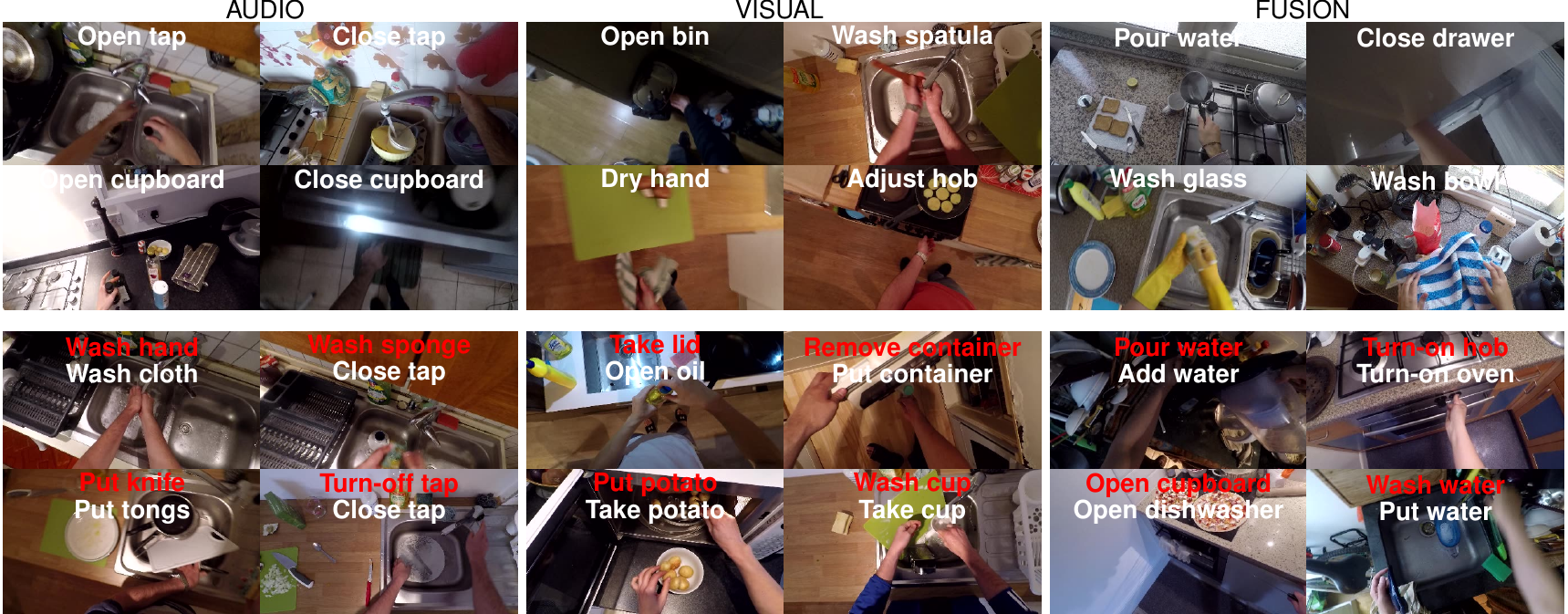}
\caption[]{Qualitative results for the unweighted multimodal \textit{action} classification experiment. The top and bottom rows shows true and false positive prediction, respectively. The columns indicate when the audio, vision, or their fusion scores were the final multimodal decision. The true and false labels are shown in white and red colors.}
\label{fig:qualitativeResults}
\end{center}
\end{figure*}

\paragraph{Action}
The results of our experiments on the home made partition are presented in the upper part of Table \ref{tab:actionRecognition}. They show that using multimodal information outperforms single audio or visual classification when training separately the \textit{verb} and \textit{noun} classifiers. Even though the best top-1 accuracy was achieved when considering the \textit{verb} and \textit{noun} labels as a single action classification problem, all other performance metrics were lower than when considering them separately. The upper part of Table \ref{tab:actionRecognition} shows that the accuracy of \textit{noun} is diminished when adding audio scores to the \textit{action} classification. The categories that changed their prediction on the multimodality setting for the action classifier are presented in Fig. \ref{fig:actionAccDiff}. 

Some qualitative results of true and false positive predictions are shown in Fig. \ref{fig:qualitativeResults}. The action predictions made by the audio modality rely more on the \textit{verb} than \textit{noun} classification. For instance, it correctly predicts the verb \textit{wash} in Fig. \ref{fig:qualitativeResults}, but falsely predicts the noun \textit{hand} instead of \textit{cloth}. The visual modality obtained similar accuracy on \textit{verb} and \textit{noun} classification. Their predictions fail in cases where the actions occur in similar contexts, for example, the actions \textit{take cup} and \textit{wash cup} can occur in the sink, as illustrated in in Fig. \ref{fig:qualitativeResults}. The combination of audio and visual modalities help to disambiguate actions where the objects are occluded such as \textit{open drawer} and \textit{close drawer}, as shown in first row of the fusion column in Fig. \ref{fig:qualitativeResults}. Their complementarity also helps on the classification of actions with the same \textit{verb}, but different \textit{noun}, such as the case of \textit{washing} \textit{glass} and \textit{bowl}, shown in the second row of the fusion column in Fig. \ref{fig:qualitativeResults}.

\begin{table*}[!t]
\vspace{-1mm}
\begin{center}
\resizebox{\textwidth}{!}{%
\begin{tabular}{lccl|ccc|ccc|ccc|ccc}
& & &%
&\multicolumn{3}{c|}{\textbf{Top-1 Accuracy}}%
&\multicolumn{3}{c|}{\textbf{Top-5 Accuracy}}%
&\multicolumn{3}{c|}{\textbf{Avg Class Precision}}%
&\multicolumn{3}{c}{\textbf{Avg Class Recall}}\\\cline{5-16}
& & &%
&VERB &NOUN &ACTION%
&VERB &NOUN &ACTION%
&VERB &NOUN &ACTION%
&VERB &NOUN &ACTION\\\cline{2-16}
\multirow{15}{*}{\rotatebox{90}{\textbf{S1}}}
& \multirow{2}{*}{\rotatebox{90}{Audio}} &%
&Traditional Dilated (Verb+Noun)%
&35.11 &10.65 &03.95%
&75.33 &28.63 &13.01%
&15.43 &06.19 &01.75%
&11.03 &06.64 &01.26\\
& &%
&Traditional Dilated (Action)%
&34.06 &05.31 &07.43%
&73.51 &18.24 &20.94%
&05.19 &02.74 &01.81%
&07.85 &04.14 &03.08\\
\cline{2-16}
& \multirow{9}{*}{\rotatebox{90}{Visual}} &%
& TSN BNInception (FUSION)~\cite{Damen2018EPICKITCHENS}%
&48.23 &36.71 &20.54%
&84.09 &62.32 &39.79%
&47.26 &35.42 &10.46%
&22.33 &30.53 &08.83\\
& &%
&TSN ResNet-50 (Verb+Noun) (FUSION)%
&55.08 &38.59 &24.38%
&86.36 &64.16 &45.37%
&43.69 &38.59 &14.91%
&28.63 &32.10 &12.12\\
& &%
&TSN ResNet-50 (Action) (FUSION)%
&38.62 &25.84 &27.95%
&79.51 &54.18 &49.12%
&10.50 &24.63 &14.13%
&14.30 &21.12 &14.61\\
& &%
& LSTA (two stream)~\cite{sudhakaran2019lsta}%
&59.55 &38.35 &30.33%
&85.77 &61.49 &49.97%
&42.72 &36.19 &14.46%
&38.12 &36.19 &17.76\\\cline{4-16}
& &%
& 3rd Place Challenge~\cite{FbkEPICChallengeSubmission2019}%
&63.34 &44.75 &35.54%
&89.01 &69.88 &57.18%
&63.21 &42.26 &19.76%
&37.77 &41.28 &21.19\\
& &%
& 2nd Place Challenge~\cite{GhadiyaramEPICChallengeSubmission2019}%
&64.14 &47.65 &35.75%
&87.64 &70.66 &54.65%
&43.64 &40.52 &18.95%
&38.31 &45.29 &21.13\\
& &%
& 1st Place Challenge~\cite{BaiduEPICChallengeSubmission2019}%
&\textbf{69.80} &\textbf{52.27} &\textbf{41.37}%
&\textbf{90.95} &\textbf{76.71} &\textbf{63.59}%
&\textbf{63.55} &\textbf{46.86} &\textbf{25.13}%
&\textbf{46.94} &\textbf{49.17} &\textbf{26.39}\\\cline{2-16}
& \multirow{6}{*}{\rotatebox{90}{Multimodal}}%
& \multirow{3}{*}{\rotatebox{90}{\footnotesize{Verb+Noun}}}%
& Ours %
&56.37 &37.69 &24.00%
&85.47 &63.45 &44.66%
&48.15 &38.02 &13.49%
&25.54& 30.31& 10.50\\
& &%
& Ours (Weighted)%
&56.44 &\textcolor{blue}{\textbf{39.42}} &25.26%
&85.87 &\textcolor{blue}{\textbf{65.27}} &46.27%
&\textcolor{blue}{\textbf{51.39}} &\textcolor{blue}{\textbf{38.36}} &14.88%
&26.66 &32.88 &11.90\\
& &%
& Ours (FC)%
&\textcolor{blue}{\textbf{58.88}} &39.13 &27.35%
&\textcolor{blue}{\textbf{87.15}} &64.83 &47.68%
&46.36 &37.92 &\textcolor{blue}{\textbf{16.63}}%
&\textcolor{blue}{\textbf{38.13}} &\textcolor{blue}{\textbf{34.90}} &15.00\\\cline{3-16}
& & \multirow{3}{*}{\rotatebox{90}{\small{Action}}}%
& Ours%
&40.80 &22.29 &28.83%
&81.04 &50.59 &49.68%
&11.43 &23.00 &15.89%
&13.40 &17.90 &14.18\\
& &%
& Ours (Weighted)%
&41.22 &24.29 &29.09%
&81.16 &53.25 &\textcolor{blue}{\textbf{50.57}}%
&11.24 &23.61 &15.35%
&14.04 &19.72 &14.51\\%
& &%
& Ours (FC)%
&44.64 &30.64 &\textcolor{blue}{\textbf{29.13}}%
&76.41 &59.39 &49.71%
&19.90 &32.28 &16.51%
&21.99 &25.28 &\textcolor{blue}{\textbf{16.54}}\\
\hline
\multirow{15}{*}{\rotatebox{90}{\textbf{S2}}} 
& \multirow{2}{*}{\rotatebox{90}{Audio}} &%
&Traditional Dilated (Verb+Noun)%
&30.73 &07.20 &02.53%
&67.26 &21.65 &09.90%
&13.92 &04.52 &01.90%
&09.79 &04.68 &01.20\\
& &%
&Traditional Dilated (Action)%
&31.96 &03.89 &03.96%
&64.73 &13.96 &13.45%
&05.05 &01.66 &00.97%
&07.91 &04.04 &01.94\\
\cline{2-16}
& \multirow{9}{*}{\rotatebox{90}{Visual}} &%
&TSN BNInception (FUSION)~\cite{Damen2018EPICKITCHENS}%
&39.40 &22.70 &10.89%
&74.29 &45.72 &25.26%
&22.54 &15.33 &05.60%
&13.06 &17.52 &05.81\\
& &
& TSN ResNet-50 (Verb+Noun) (FUSION)%
&45.72 &24.89 &14.95%
&77.06 &49.37 &31.07%
&24.44 &20.30 &08.79%
&18.04 &18.96 &10.10\\
& &
& TSN ResNet-50 (Action) (FUSION)%
&36.63 &18.06 &17.14%
&75.28 &42.03 &34.65%
&11.38 &10.89 &07.27%
&12.96 &14.38 &10.78\\
& &
&LSTA (two stream)~\cite{sudhakaran2019lsta}%
&47.32 &22.16 &16.63%
&77.02 &43.15 &30.93%
&31.57 &17.91 &08.97%
&26.17 &17.80 &11.92\\\cline{4-16}
& &%
& 3rd Place Challenge~\cite{FbkEPICChallengeSubmission2019}%
&49.37 &27.11 &20.25%
&77.50 &51.96 &37.56%
&31.09 &21.06 &09.18%
&18.73 &21.88 &14.23\\
& &%
& 2nd Place Challenge~\cite{GhadiyaramEPICChallengeSubmission2019}%
&55.24 &33.87 &23.93%
&80.23 &58.25 &40.15%
&25.71 &28.19 &\textbf{15.72}%
&25.69 &29.51 &17.06\\
& &%
& 1st Place Challenge~\cite{BaiduEPICChallengeSubmission2019}%
&\textbf{59.68} &\textbf{34.14} &\textbf{25.06}%
&\textbf{82.69} &\textbf{62.38} &\textbf{45.95}%
&\textbf{37.20} &\textbf{29.14} &15.44%
&\textbf{29.81} &\textbf{30.48} &\textbf{18.67}\\\cline{2-16}
& \multirow{6}{*}{\rotatebox{90}{Multimodal}}%
& \multirow{3}{*}{\rotatebox{90}{\footnotesize{Verb+Noun}}}%
& Ours%
&46.88 &25.16 &14.58%
&77.13 &48.69 &31.00%
&\textcolor{blue}{\textbf{28.72}} &\textcolor{blue}{\textbf{16.63}} &08.93%
&17.25 &17.83 &08.55\\
& &
& Ours Weighted%
&47.46 &25.95 &15.74%
&\textcolor{blue}{\textbf{77.16}} &\textcolor{blue}{\textbf{50.12}} &31.85%
&28.71 &16.47 &09.26%
&17.85 &19.21 &09.94\\
& &
& Ours FC%
&\textcolor{blue}{\textbf{47.49}} &\textcolor{blue}{\textbf{26.36}} &15.98%
&76.68 &49.37 &31.75%
&24.64 &20.61 &\textcolor{blue}{\textbf{09.80}}%
&\textcolor{blue}{\textbf{20.59}} &\textcolor{blue}{\textbf{21.35}} &10.03\\\cline{3-16}
& & \multirow{3}{*}{\rotatebox{90}{\small{Action}}}%
& Ours%
&38.37 &15.23 &\textcolor{blue}{\textbf{18.40}}%
&75.15 &39.84 &35.64%
&10.93 &11.60 &06.88%
&11.75 &13.31 &10.91\\
& &%
& Ours (Weighted)%
&38.10 &16.76 &18.23%
&75.38 &42.23 &\textcolor{blue}{\textbf{35.68}}%
&11.58 &12.83 &07.72%
&11.79 &14.16 &11.26\\
& &%
& Ours (FC)%
&40.87 &20.38 &17.65%
&69.27 &45.82 &33.73%
&15.72 &15.35 &09.61%
&17.34 &16.95 &\textcolor{blue}{\textbf{12.20}}\\%
\end{tabular}}
\caption{Performance comparison with EPIC Kitchens challenge baseline results. The results highlighted in \textcolor{blue}{\textbf{bold blue}} are the best obtained by our method.}
\label{tab:epicKitchensChallenge}
\end{center}
\end{table*}

\paragraph{Comparison with other methods}

Table \ref{tab:epicKitchensChallenge} shows the results obtained on the EPIC Kitchen Challenge board using the models trained on the home made partition. These results indicate that directly training over the \textit{action} performs better than the combination of \textit{verb+noun}. Our multimodal models obtained better scores than the challenge baseline and has similar results as previous works \cite{sudhakaran2019lsta}. Additionally, the results obtained on the unseen participants (S2) test split are in the top-ten ranking of the first challenge. We can also observed that the unweighted addition as fusion method for $noun$ diminishes the aggregate and per-class performance, not only for our method but also in the baseline results \cite{Damen2018EPICKITCHENS}.

The results on the comparison data split originally presented on \cite{Baradel_2018_ECCV} are shown in Table \ref{tab:actionRecognitionBaradel}. Our method obtained an improvement in accuracy of 5.18\% with respect to the ORN method \cite{Baradel_2018_ECCV}. This test on unseen kitchens showed that methods that only rely on RGB underperform optical flow methods. Likewise, they also showed that audio classification methods can have similar performance to visual methods. The addition of auditory sources increased the performance of the best visual method by 3.47\%.

\paragraph{Comparison of audio CNN architectures}

The VGG-11 and the Traditional Dilated network have similar classification performance. The results on Table \ref{tab:actionRecognition} indicate that the Traditional Dilated network has better results on seen test users, but the results on Table \ref{tab:actionRecognitionBaradel} shows that the VGG-11 network outperforms the Traditional Dilated network on unseen test users.

\subsection{Discussion}

The experimental results show that our model improves the top-1 accuracy by 3.61\% in average for \textit{verb}, \textit{noun}, and \textit{action} classification on our home made data partition. Likewise, the results suggest that audiovisual multimodality benefits the classification of \textit{verb}, and consequently \textit{action}, more than for the classification of \textit{noun}. Furthermore, although multimodality improves the aggregate performance metrics and the avg. class precision for $noun$ classification, the unweighted fusion decreases their value as observed in Table \ref{tab:actionRecognition}. This might be as a consequence of three main reasons. First, an interacting object can produce several sounds and not being described by one in particular. For instance, the sounds of a fridge being opened and closed are characteristically different. Second, rather than describing an object, sounds are better suited for describing the materials they are made of. For example, water and milk are liquids and their emitting sound while being poured is indistinguishable. Third, objects lack of a time dimension, but doing an action and making a sound involve time. Nonetheless, not all actions produce any sound, like checking the coffee pot. The verb confusion matrix in Fig. \ref{fig:verbConfusionMatrix} shows that the multimodality classification on verbs fails on categories that does not produce any sound and that are more visually abstract, like \textit{checking} the heat. Additionally, harder audiovisual classes are \textit{empty}, \textit{flip}, and \textit{squeeze}, as seen in Fig.\; \ref{fig:verbConfusionMatrix} and Fig.\; \ref{fig:verbAccDiff}. The most discriminative input source for the \textit{noun}, \textit{verb}, and \textit{action} comes from the RGB frames as observed from Tables \ref{tab:actionRecognition} and \ref{tab:epicKitchensChallenge}. However their performance decreases when the test is performed on images from unseen persons and the optical flow achieves a higher accuracy, as seen in Table \ref{tab:actionRecognitionBaradel}. 

\section{Conclusions}
\label{sec:fine}
We presented a multimodal approach for egocentric action classification and we validated it on the EPIC Kitchens dataset. Our approach combines audiovisual input sources. Specifically, its audio input is the spectrogram extracted from the raw audio of the video, while its visual inputs are RGB and optical flow frames. We tested and analyzed our approach for classifying each separate category (\textit{verb} and \textit{noun}) or merged (\textit{action}). The obtained results show that our model improves the top-1 accuracy by 3.61\% in average. Additionally, the results suggest that multimodal information is spacially beneficial for the \textit{verb} recognition problem. Indeed, our multimodal approach outperformed the state of the art methods on \textit{verb} classification by 5.18\% accuracy.

\footnotesize{
\textbf{Acknowledgment}          
This work was partially funded by TIN2018-095232-B-C21, 2017 SGR 1742, Nestore, Validithi, 20141510 (La MaratoTV3) and CERCA Programme/Generalitat de Catalunya. A.C. is supported by a doctoral fellowship from the Mexican Council of Science and Technology (CONACYT) (grant-no. 366596). We acknowledge the support of NVIDIA Corporation with the donation of Titan Xp GPUs.}
{\small
\bibliographystyle{ieee_fullname}

}

\end{document}